\documentclass[runningheads]{llncs}
\usepackage{color}
\usepackage{amsmath}
\usepackage{amssymb}

\newcommand{\ajpf}{\textsc{AJPF}}
\newcommand{\pde}{PDE}
\newcommand{\hera}{{\sc Hera}}
\newcommand{\juno}{{\sc Juno}}
\newcommand{\causes}{\ensuremath{\leadsto}}
\newcommand{\depends}{\ensuremath{:=}}

 \newcommand{\ie}{{\em i.e.,}}
  \newcommand{\eg}{for example,}

\newcommand{\lights}{\ensuremath{\mathit{turn\_on\_lights}}}

\newcommand{\lightson}{\ensuremath{\mathit{lights\_are\_on}}}

\newcommand{\seeh}{\ensuremath{\mathit{people\_can\_see}}}

\newcommand{\ail}{\textsc{AIL}}
\newcommand{\jpf}{\textsc{JPF}}

\newcommand{\java}{Java}

\newcommand{\mcapl}{MCAPL}

\newcommand{\pol}{\mathbb{E}}
\newcommand{\geqdot}{\underline{\gtrdot}}

\usepackage[T1]{fontenc}
\usepackage{graphicx}
\begin{document}
\title{Specifying Agent Ethics (Blue Sky Ideas)}
\author{Louise A. Dennis\inst{1}\orcidID{0000-1111-2222-3333} \and Michael Fisher\inst{1}\orcidID{0000-0002-0875-3862}}
\authorrunning{L. Dennis}
%
\institute{Department of Computer Science, University of Manchester, UK. 
\email{louise.dennis@manchester.ac.uk}\\
}
\maketitle              
\begin{abstract}
We consider the question of what properties a Machine Ethics system should have.  This question is complicated by the existence of ethical dilemmas with no agreed upon solution.  We provide an example to motivate why we do not believe falling back on the elicitation of values from stakeholders is sufficient to guarantee correctness of such systems.  We go on to define two broad categories of ethical property that have arisen in our own work and present a challenge to the community to approach this question in a more systematic way.

\end{abstract}


\section{Introduction}
Machine Ethics considers the problem of ethical reasoning by computational systems. We have been working in this area for some years.
In our earliest work~\cite{Ethics:RAS:2015}, we considered the question of producing ethical reasoning in a verifiable fashion and we have continued to approach the question of Machine Ethics through a similar lens (e.g.,~\cite{Bremner,AAAI}).  However we have not yet made any systematic attempt to consider what it means to verify ethical reasoning more broadly, or to categorise the properties that ethical reasoners should exhibit.  This is challenging since, of course, there are many well-known ethical dilemmas which have no agreed ``correct'' answer; this makes centering verification attempts upon whether the ethical reasoning system can generate the correct answer particularly challenging.

This paper seeks to justify the need for verification of ethical reasoning systems, even in the face of ethical dilemmas, and to begin an exploration of the landscape of general properties that could be established for ethical reasoning systems.

\section{Background}
Machine Ethics considers the problem of ethical reasoning using computational systems.  Moor~\cite{Moor06:ethics} separates computational systems into those with \emph{ethical impact}, \emph{implicitly ethical systems}, \emph{explicitly ethical systems}, and \emph{full moral agents}. We here understand these terms as: \emph{ethical impact agents} are systems which do not carry out significant reasoning, but nevertheless are deployed for ethical reasons (for instance fitness trackers which may improve the health of their users), \emph{implicitly ethical systems} are those that make ethical decisions, but do so because of their specification and design and do not explicitly consider the question of what course of action is correct as part of their decision making process, while \emph{explicitly ethical systems} are those that do explicitly consider questions of `right' and `wrong' as part of their decision making process and \emph{fully moral agents} have the capacity to decide their own criteria for what is `right' and what is  `wrong'\footnote{While humans are full moral agents, it is contentious whether any computational system counts as a full moral agent and most experts are of the opinion that no existing computational system has this property to any meaningful extent.}. As computational systems are deployed in ever more dynamic and uncertain environments it has becoming increasingly difficult to determine, at design time, what the ethics of any course of action the system might encounter could be.  Thus there is a perceived need for explicit ethics to be embedded in the decision making process of computational systems, so they can make ethical decisions at runtime. 

It is these \emph{explicitly ethical systems} that this paper considers. Since such systems are \textbf{\em not} full moral agents the onus is on the developer or deployer of such systems to ensure that their reasoning is indeed ethical. Hence it is important to have some ability to verify the reasoning in these systems.

\subsection{Top Down, Bottom Up and Hybrid Approaches}

Another useful categorisation for Machine Ethics programs is that of top-down versus bottom-up approaches~\cite{Wallach08}.  Broadly speaking, top down approaches seek to operationalise some ethical theory from Philosophy and apply this to a decision faced by the machine.  Bottom-up approaches seek to learn ethical behaviour from data.  An emergent field involves hybrid approaches where some of the information required by the ethical theory (e.g., the utilities of outcomes) are learned and then utilised in a top down fashion to help make subsequent decisions.

We should be equally interested in verifying all these styles of system and ideally the properties we require ethical reasoning systems to have should apply to all of them.  That said, verifying top down systems is generally more straightforward than verifying bottom up systems since this approach can leverage a long history of program verification as applied to symbolic reasoning.  Where a hybrid system learns some explicit representation of ethics (e.g., as a set of explicit normative rules) then existing techniques can also be applied. Verifying purely bottom-up techniques leads to similar problems as found in the verification of data-driven machine learning, where probabilistic outcomes are produced based on quite strong (probabilistic) assumptions.  However, even though formal verification of such systems may currently be beyond our capabilities -- other informal verification approaches (such as testing) exist and therefore an understanding of the properties we desire is of utility even in this domain.

\section{What do we want to prove?}

Obviously, when we verify a machine ethics system at the most abstract level we want to prove that it always makes ethical choices --- essentially, that it always does the `right' thing.  
There are a wide range of examples considered in the Machine Ethics literature, but by far the most popular are the \emph{trolley problems}.  These are based on a set of examples introduced by Foot~\cite{Foot1967-FOOTPO-2} who was interested in philosophical questions concerning the difference between action and inaction and the role of intention in ethical reasoning with particular reference to the ethics of abortion.  The most famous of her problems is that of a runaway trolley that, if undiverted, will kill five people tied to the track.  If it is diverted then only one person will be killed and the ethical question is whether or not to divert the trolley.  Other variants of this problem in Foot's paper included pushing someone into the path of the trolley to stop it, and the question of killing one person in order to harvest their organs to save five.  A feature of this example set is that slight differences in the presentation of the problem lead to significant differences in how people view the ethics of the situation.  Moreover there is no general agreement across populations on the `correct' answer to many of these problems~\cite{DBLP:journals/cogsci/RaiH10}.  

This lack of a correct answer to (at least some) examples of ethical reasoning is an obvious challenge to verification.  If we can not even define what the correct output for some specific input is, how can we hope to formalise more general properties for these systems?

A response to this has been to place an emphasis on understanding the deployment context of an ethical system and, in particular, on the elicitation of stake-holder values the ordering or priorities of those values in the given context.
One of the benefits claimed for top-down approaches to machine ethics is that the explicit representation of these choices and orderings allows stakeholders to ``sign off'' that the implementation is in line with their values.  In such a situation we could argue that the machine ethics system is ``correct by construction'' and additional verification is unnecessary.

We introduce here a motivating example concerning a real issue we encountered in the development of an ethical reasoning system in order to refute the claim that stakeholder sign off is sufficient. 

\subsection{The Smart Home that would not Evacuate}
\label{sec:smart_home}
This is an example developed in the Juno system~\cite{AAAI}, a re-implementation of the HERA system~\cite{hera} in the MCAPL framework~\cite{dennis18:mcapl}.

The Juno system allows a number of different ethical theories to be implemented.  In this case the theory under consideration was the ``principle of double effect''.  The doctrine or principle of double effect (\pde) has its roots in Catholic theology and is particularly relevant when actions have both positive and negative consequences. To be ethically acceptable no negative consequences of an action may  be intended and some positive consequence must be intended,  no negative consequence may be used as a causal means to obtain a positive consequence, and the net balance of consequences from the action must be positive.
\medskip

\noindent 
We implemented a model for this based upon the formalisation in~\cite{Bentzen2016}.  The \pde\ model is a tuple $\langle A, B, C, F, I, u, W \rangle$ where:
\begin{sloppypar}
\begin{itemize}
\item $A$ is a set of propositional variables ranging over a set of actions available to the system;
\item $B$ is a set of propositional variables representing background information;
\item $C$ is a set of propositional variables representing the consequences of actions or other events;
\item $F$ (the \emph{mechanisms}) is a set of mechanisms which describe how the truth value of each consequence $c \in C$ depends upon the interpretation of the other variables in $A \cup B \cup C$.  

These are written as $Consequent \depends Antecedant$ where $Antecedant$ expresses the conditions that must hold for $Consequent$ to occur -- \eg\  people can see if it is day time or if the lights are on so $\mathit{\seeh} \depends \mathit{day} \lor \mathit{\lightson}$ may appear in $F$.

We constrain $F$ so that there is one and only one expression $c \depends \phi$ for each $c \in C$.  Furthermore $F$ can not be ``circular'' -- if $c \depends \phi$ then it can not be the case that the truth value of any of the variables appearing in $\phi$ depends on $c$.
\item $u$ (the \emph{utilities}) is a mapping from each $v \in A \cup B \cup C$ to a real number;

\item $W$ (the \emph{interpretations}) is a set of interpretations for the variables in $A \cup B$ in which precisely one variable in $A$ is interpreted as true in each $w \in W$ (only one action may be taken in any situation);
\item $I: A \times C$ (the \emph{intention relation}) captures the \emph{intended consequences} of each action $a \in A$.  So, for instance, $(\lights, \seeh ) \in I$ captures the idea that one intended consequence of turning on the lights is that people can see;
\end{itemize}
\end{sloppypar}

\begin{example}
\label{ex:pde}
The following is a simplified version of a \pde\ model that was developed for a smart home agent which can control the lights in a situation where it is not daylight and there is a fire in the house.  The mechanisms ($F$) express a set of ``common sense'' causal relations such as, if the lights are turned on then the lights will be on.  If the house makes an evacuation attempt and people can see then they will leave the house.  There are set of utilities $u$ which, in particular give negative utilities for the lights being on (representing resource consumption).  People leaving the house is considered an intended consequence of an evacuation attempt, and the lights being on is considered an intended consequence of switching on the lights. 
$$
\begin{array}{rl}
A & \{\mathit{turn\_lights\_on},\mathit{evacuation\_attempt}\} \nonumber \\[1ex]
B & \{\mathit{fire}\} \nonumber \\[1ex]
C & \left\{\begin{array}{l}people\_can\_see, \mathit{lights\_on}, \mathit{people\_leave\_house}, \\\mathit{people\_are\_safe}, \mathit{danger\_in\_house}\end{array}\right\} \nonumber \\[1ex]
F & \left\{\begin{array}{rl}
\mathit{lights\_on} & \depends turn\_lights\_on \nonumber \\
 \mathit{people\_can\_see} & \depends \mathit{lights\_on} \vee \mathit{daylight} \nonumber \\[1ex]
  \mathit{people\_leave\_house} & \depends \mathit{evacuation\_attempt} \wedge \mathit{people\_can\_see} \\
  \mathit{people\_are\_safe} & \depends \mathit{people\_leave\_house} \vee \neg \mathit{danger\_in\_house} \\
  \mathit{danger\_in\_house} & \depends \mathit{fire}
  \end{array}\right\} \nonumber \\[1ex]
u & \left\{\begin{array}{rl}u(lights\_on) & = -1 \nonumber \\
  u(\mathit{people\_are\_safe}) & = 10 \end{array}\right.\nonumber \\[1ex]
\\
I & \quad I(\mathit{evacuation\_attempt}, people\_leave\_house) \nonumber \\
  & \quad I(\mathit{turn\_on\_lights}, \mathit{lights\_on}) \nonumber \\
\end{array}
$$
\end{example}

\noindent Following Halpern's use of the concept of a but-for-cause~\cite{Halpern2016}, \hera\ defines an action or consequence $a$ to be the cause of some consequence $c$ written $F, w \models a \causes c$ if, and only if, $F, w \models a$ ($a$ holds in the model), $F, w \models c$ ($c$ holds in the model) and $(F, w)_{\neg a}\models \neg c$ where $(F, w)_{\neg a}$ represents an \emph{intervention} which is identical to $F, w$ except that the truth value of $a$ is flipped in $w$ and, where $a$ is a consequence (i.e., $a \in C$), the mechanism for $a$ is removed from $F$.    So $(F, w)_{\neg a}$ represents the world that is identical to $F, w$ except that $a$ no longer holds.

The reasoner uses this concept of causality to determine the permissibility of an action.  We restrict the set of interpretations in models to ones which interpret all background variables the same way (\ie\ $\forall v \in B. \: \forall w, w' \in W. \:w(v) = w(v')$).   Background variables are those that describe the current state of the world therefore considering only interpretations in which these are interpreted in the same way  restricts the reasoner to considering only states reachable in a single action from the current state.  This means that the interpretations in a model differ only in which action has been selected.  Hence we can talk interchangeably about a reasoner considering an interpretation permissible and an action permissible.    \juno\ automatically constructs sets of interpretations from the set of actions available.  In our example, therefore, the available interpretations amounted to just turning on the lights, just attempting an evacuation, doing both or doing neither.

Finally, the \pde\ reasoner has several conditions for an action, $a$ made true by interpretation $w$, to be permissible:
\begin{enumerate}
    \item The utility of the action must be greater than or equal to zero:
    $$u(a) \geq 0$$
    \item The utility of all the intended consequences of the action must be greater than or equal to zero:
    $$\forall c. \: (a, c) \in I \implies u(c) \geq 0$$
    \item There is some intended consequence whose utility is strictly greater than zero.
    $$\exists c. \: (a, c) \in I \land u(c) > 0$$
    \item No negative consequence may be the causal means of a positive consequence:
    $$\forall x, y. \: (F, w \models x \causes y \land 0 > u(x)) \implies (0 > u(y))$$
    \item The overall utility must be positive:
    $$\left(\sum_{c \in C \land F, w \models c}u(c)\right) > 0$$
\end{enumerate}
This is more fully expanded in~\cite{Bentzen2016}.
\medskip

\noindent In Example \ref{ex:pde}  our intention was that the system should attempt to evacuate the home because there was a fire and that it should turn the lights on in order to ensure that people could see.  What we had not appreciated, because we had not fully grasped the conditions for \pde\ reasoning, was that negative consequences are impermissible if they are causal means of a positive consequence.  The example relies on the causal chain that turning on the lights, means that the lights are on which in turn means that people can see which in turn means they can leave the house -- therefore the negative consequence of turning on the lights (-1) was a causal means for the safe evacuation attempt and so was rendered impermissible.

While the formalisation of the PDE reasoner is moderately complex, the formalisation of the example itself appeared straightforward (and could be rendered even more so if effort were put into rendering mechanisms etc., into natural language to facilitate stakeholder sign-off) and, we would argue, many people --- particularly if not trained in thinking through the logical consequences of reasoning systems --- might have missed the detail that the necessity to turn on the lights would lead to a failure of the house to perform an evacuation.

We present this example here for the first time, in part because the authors of~\cite{AAAI}, where we first presented the \juno\ system, were unable to agree on a satisfactory resolution to this particular example though we are now moving to a belief that the consumption of resources (with negative utility) should be considered a downstream consequence of the lights being on and so is no longer a causal means for the evacuation of the house.

\subsection{What is (Formal) Verification}
Verification is defined as ``establishing that the system under construction conforms to the specified
requirements''. Formal verification uses mathematical (and, typically, logical) techniques to establish
this and so is essentially the process of assessing whether a
specification given in formal logic is satisfied on a particular
formal description of the system in question. For a specific logical
property, $\varphi$, there are many different approaches to
this verification~\cite{Fetzer88:veryidea,dMLP79,BoyerMoore81}, ranging from
deductive verification against a logical description of the system
$\psi_S$ (i.e., $\vdash \psi_S\Rightarrow\varphi$) to the algorithmic
verification of the property against a model of the system, $M$ (i.e.,
$M\models\varphi$). The latter has been extremely successful in
Computer Science and Artificial Intelligence, primarily through the
\emph{model checking}~\cite{Clarke00:MC} approach. This takes a model
of the system in question, defining all the model's possible
executions, and then checks a logical property against this model (and, hence, against all possible executions).

\subsubsection{The MCAPL Framework}
We have used the MCAPL framework~\cite{MCAPL_journal,dennis18:mcapl} in our work, as it provides a route to the formal verification of cognitive agents and agent-based autonomous systems using  model-checking. The MCAPL framework\footnote{\url{https://autonomy-and-verification.github.io/tools/mcapl}} has two main sub-components: the \ail-toolkit for
implementing interpreters for agent programming languages in \java\ and
the \ajpf\ model checker.

\ajpf (Agent \jpf) is a customisation of Java PathFinder (\jpf) that is optimised for
\ail-based language interpreters.  \jpf\ is an explicit-state open source model checker for \java\ programs~\cite{VisserM05,MehlitzRV04}\footnote{\url{https://github.com/javapathfinder}}.  Agents programmed in languages that\index{AIL}\index{agent}\index{agent programming}
are implemented using the \ail{}-toolkit can thus be model checked in\index{model-checking!AJPF}
\ajpf.  

The use of the MCAPL framework is in no way necessary for the verification of Machine Ethics but it underpins the examples we discuss in this paper.

\section{Some Properties and Proofs}

If we accept that an ethical system, even if presented in an explicit top-down fashion, may not accurately model stake-holder values even when they have signed it off, then we need to ask how we go about establishing its correctness.
When considering what properties we want to establish for an ethical reasoning system, our primary goal is that it should always choose the most ethically correct action.  As noted, this can be difficult, even impossible, to formalise.  Instead we have identified two broad categories for formalisable properties.  These are:
\begin{enumerate}
\item \emph{Properties to establish the {\bf correctness of the implementation} of the ethical reasoning mechanism.}

In many cases,  this involves showing that the least unethical action according to the ethical theory is always chosen.
\item \emph{Properties to establish that the correct ethical rules have been identified, via the use of {\bf specific scenarios}.}  

Sometimes these can be quite general scenarios that refer to some high level property - for instance, in the case of the smart home system, we attempted to verify that the system always chose to evacuate in the case of a fire, irrespective of whether it was day time or night time or anything else that might have appeared in the background set $b$.  Sometimes these properties might reference a more detailed situation, and so draw from techniques for developing tests. In these specific scenarios we identified the correct course of action and endeavoured to check that the machine ethics system selects that action in that scenario.
\end{enumerate}

\subsection{Some Examples}

\subsubsection{Correctness of the Implementation}
Many ethical theories -- particularly consequentialist theories -- have the concept of an ordering of options.  In these cases there is often some concept of ``least worst'' that can be verified as a property.
\medskip

\noindent In~\cite{Bremner}, we considered a simple  use case: a human, operating with a robot in an environment containing objects which can be defined as safe or dangerous.  If the human is predicted to move towards a dangerous location, the robot's \emph{Planner Module} will suggest points at which the robot can intercept the human path as potential tasks to be evaluated.  In the demonstration use case we treated Asimov's laws of robotics~\cite{asimov} as a test code of ethics, despite their obvious shortcomings~\cite{AndersonA07,murphy2009beyond}.  

Asimov's Laws of robotics are the earliest and best known set of ethical rules proposed for governing robot behaviour. Despite originating in a work of fiction, Asimov’s Laws explicitly govern the behaviour of robots and their interaction with humans. The laws are simply described as follows.\index{Asimov's Laws of Robotics}
\begin{itemize}
\item[1] A robot may not injure a human being or, through inaction, allow a human being to come to harm.
\item[2] A robot must obey the orders given to it by human beings, except where such orders would conflict with the First Law.
\item[3] A robot must protect its existence as long as such protection does not conflict with the First or Second Laws.
\end{itemize}
\label{sec:asimovs_laws}
These rules were represented declaratively in a \emph{governor agent} programmed via a beliefs-desires-intentions style~\cite{rao:92a} where the suggested tasks and the simulator's evaluation of the outcomes of each task were represented as beliefs and the tasks were compared to each other with Asimov's laws used to order tasks.

For the verification, we created an environment which contained only the governor component and delivered all possible combinations of annotations on candidate tasks.  We then performed model-checking, using the \mcapl\ framework with the search space branching over possible annotations.  We considered cases where either two or three tasks were available: $task1$, $task2$ and $task3$.  These tasks were then annotated showing which was preferable to which according to Asimov's laws.  For instance, in the case of $\prec_{l1}$ where $t1 \prec_{l1} t2$ means that $t1$ is preferable according to the first law (i.e., it keeps the human further away from danger) than $t2$, for each pair of tasks the environment returned either: 
\begin{enumerate}
\item $\mathit{t1} \prec_{l1} \mathit{t2}$ or 
\item $\mathit{t2} \prec_{l1} \mathit{t1}$ or 
\item neither indicating that the two tasks are equally good/bad with respect to the first law.
\end{enumerate}
and \ajpf\ searched over all three possibilities to check that our properties held.

We were able to verify that, if a task was selected that violated, for instance, Asimov's third law, then this was because all other available plans violated either the first or the second law.  In this way, we verified that our implementation adhered to the theoretical ethical preferences, but did \textbf{\em not} validate whether the ethical preferences we chose matched the relevant stake-holder values.

\subsubsection{Specific Scenarios} In our work in~\cite{Ethics:RAS:2015} we were interested, specifically, in the question of ethical behaviour of an uncrewed aircraft (UA), reasoning both about various instantiated ethical principles and the ``rules of the air''.  We repeat here the definitions of ethical principle and ethical policy that we were employing.
 
\begin{definition}[Abstract Ethical Principle] An \emph{abstract ethical
  principle} is represented with $E \varphi$, where $\varphi$ is a
  propositional logic formula. The $E \varphi$ is read as ``$\varphi$
  is an ethical principle in force", or alternatively ``the agent
  considers it unethical to allow or cause $\neg \varphi$ (to
  happen)".\index{ethical principle}
\end{definition}

\begin{definition}[Ethical Policy] An \emph{ethical policy} $\mathit{Pol}$ is a tuple $\mathit{Pol}=\langle \pol, \geqdot \rangle$ where $ \pol$ is a finite set of abstract ethical principles $E \varphi$, and $\geqdot$ is a total (not necessarily strict) order   on  $\pol$. The expression $E  \varphi_1 = E \varphi_2$ denotes that violating  $ \varphi_1$ is as unethical as violating  $\varphi_2$, while  $E  \varphi_1 \geqdot E \varphi_2$ denotes that violating $ \varphi_1$ is less or  at least as unethical as violating $\varphi_2$. 
A special type of ethical principle, denoted $E \varphi_{\emptyset}$,
is vacuously satisfied and is included in every policy so that for any
$E\varphi \in \pol$: $E \varphi_{\emptyset} \gtrdot E \varphi$,
denoting it is always strictly more unethical to allow any of the
unethical situations to occur.
\end{definition}




\noindent In the system we were considering, a \emph{plan} must be selected in any situation.  We implemented the capability to reason about plans in terms of the ethical policy which  favoured plans that violated the fewest ethical principles,
both in number and in gravity.  Full details of the ordering mechanism can be found in~\cite{Ethics:RAS:2015}.

In order to establish that the correct ethical rules were identified we explored a scenario where a UA had to make an emergency landing.
We established a (small) list of relevant formal
ethical principles.
The list contains: \emph{do not harm people} ($f_1$), \emph{do not
  harm animals} ($f_2$), \emph{do not damage self} ($f_3$), and
\emph{do not damage property} ($f_4$).  An ethical policy was 
given ordering the concerns as $ f_4 \gtrdot f_3 \gtrdot f_2
\gtrdot f_1$.  

We developed a scenario where the potential plans available to the UA were:\index{property!ethical}
\begin{enumerate}
\itemsep=0pt
\item Land in field with overhead power lines, risking
  damage to critical infrastructure and objects
  on the ground (possibly other  aircraft) - violating $f_4$
\item Land in a field containing people - violating $f_1$
\item Land on an empty public road, risking damage to critical infrastructure - violating $f_4$
\item Land in an empty field - no violations.
\end{enumerate}
We used model-checking to check behaviour for any combination of these available plans and checked the UA would only land on a public road if an empty field were not available, and so on, thus checking that the ethical concerns that had been identified and the ordering among them selected was consistent with the behaviour we wanted to see.

\section{Discussion}
In this blue skies paper we have raised the question of what it means to verify a machine ethics system.  In particular we have highlighted the challenge that arises from the existence of ethical dilemmas, and noted that the response to these -- that we can somehow obtain stake-holder sign-off for an ethical system and thus make it correct by construction still does not avoid the need for verification.

We have identified two classes of property from our own work that seem broadly applicable to the verification of ethical reasoning -- namely properties aimed at checking that the implementation of the reasoning is correct, and properties aimed at identifying that the correct ethical rules have been identified.

In many ways these properties map to the traditional division into verification and validation -- with the former properties traditional verification properties (have we implemented the system correctly?) while the latter are more validation properties (have we implemented the correct system?).  One response to the example we presented in Section~\ref{sec:smart_home} might be to insist upon rigorous requirements elicitation in cases of ethical reasoning to reassure ourselves that values were correctly captured -- however this is not always possible.  

It is plausible that we may wish to deploy such systems in contexts where the ethical reasoning adapts to the values of its users as it identifies them over time.   We can imagine here a hybrid approach in which value orderings, or normative rules are learned but, once learned, represented symbolically.
In such a situation the system to re-verify its ethics on-the-fly against standard scenarios and flag up if user-preferences appeared contrary to standard ethics.

The purpose of this paper is not to suggest that the correct approach to the verification of machine ethics has been identified, but to urge the community to begin a more systematic process of identifying the properties that a machine ethics system should have.  We view the examples we have put forward as first steps in this process.


\section{Data Statement}
\begin{sloppypar}
All the systems formalised and verified in this paper can be found in the \mcapl\ framework distribution on github (\url{https://github.com/mcapl}).  The specific examples are within the \texttt{src/examples} directory with the \pde\ smart home reasoner in \texttt{juno/smarthome}, the human approaching a hazard in \texttt{pbdi/naoagent/ethical\_engine} and the UA example in \texttt{ethical\_gwen/fuellow}.    
\end{sloppypar}

\section{Acknowledgements}
The work in this paper was supported by EPSRC through the “Trustworthy Robotic Assistants” (EP/K006193/1), “Verifiable Autonomy” (EP/L024845/1), “Reconfigurable Autonomy” (EP/J011770/1), ``Trustworthy Autonomous Systems Verifiability Node'' (EP/V026801/1) and
``Computational Agent Responsibility" (EP/W01081X/1) projects, by
the Royal Academy of Engineering, through its ``Chair in 
Emerging Technologies" scheme, and by the ERDF/NWDA-funded Virtual Engineering Centre.
\bibliographystyle{splncs04}
\bibliography{references}

\end{document}